\documentclass{article}
\usepackage{spconf,amsmath,graphicx,hyperref}
\usepackage{booktabs}
\usepackage{caption}
\usepackage{comment}
\usepackage{amssymb}
\usepackage{xcolor}

\def\ie{\emph{i.e.,}}

\newcommand*{\affmark}[1][*]{\textsuperscript{#1}}

\title{Codebook-based Adaptive Feature Compression with Semantic Enhancement for Edge-Cloud Systems}

\name{Xinyu Wang\affmark[1,2], Zikun Zhou\affmark[1,2*], Yingjian Li\affmark[2], Xin An\affmark[1], Hongpeng Wang\affmark[1,2*] \thanks{* Corresponding Authors. Email: wanghp@hit.edu.cn}}
\address{\affmark[1]Harbin Institute of Technology, Shenzhen \quad \affmark[2]Pengcheng Laboratory} 
\begin{document}
%\ninept
%
\maketitle
\begin{abstract}
Coding images for machines with minimal bitrate and strong analysis performance is key to effective edge-cloud systems. Several approaches deploy an image codec and perform analysis on the reconstructed image. Other methods compress intermediate features using entropy models and subsequently perform analysis on the decoded features. Nevertheless, these methods both perform poorly under low-bitrate conditions, as they retain many redundant details or learn over-concentrated symbol distributions. In this paper, we propose a Codebook-based Adaptive Feature Compression framework with Semantic Enhancement, named \textit{CAFC-SE}. It maps continuous visual features to discrete indices with a codebook at the edge via Vector Quantization (VQ) and selectively transmits them to the cloud. The VQ operation that projects feature vectors onto the nearest visual primitives enables us to preserve more informative visual patterns under low-bitrate conditions. Hence, CAFC-SE is less vulnerable to low-bitrate conditions. Extensive experiments demonstrate the superiority of our method in terms of rate and accuracy.

\end{abstract}
\section{Introduction}
\label{sec:intro}

The rapid expansion of edge devices, coupled with increasing visual data, necessitates the deployment of deep models directly on edge platforms for efficient data processing. However, the limited computational resources of edge devices prevent them from hosting powerful, large-scale deep models. The edge-cloud system~\cite{shi2016edge,kang2017neurosurgeon,chen2019toward}, an advanced computing framework, offers a promising solution to this challenge. In this system, lightweight models are deployed on edge devices to preprocess and compress visual data before transmission to the cloud, while heavyweight models are deployed on the cloud to conduct complex inference. The key to an efficient edge-cloud system is to simultaneously reduce transmission bitrate and maintain high cloud performance.

\begin{figure}
    \centering
    \includegraphics[width=1\linewidth]{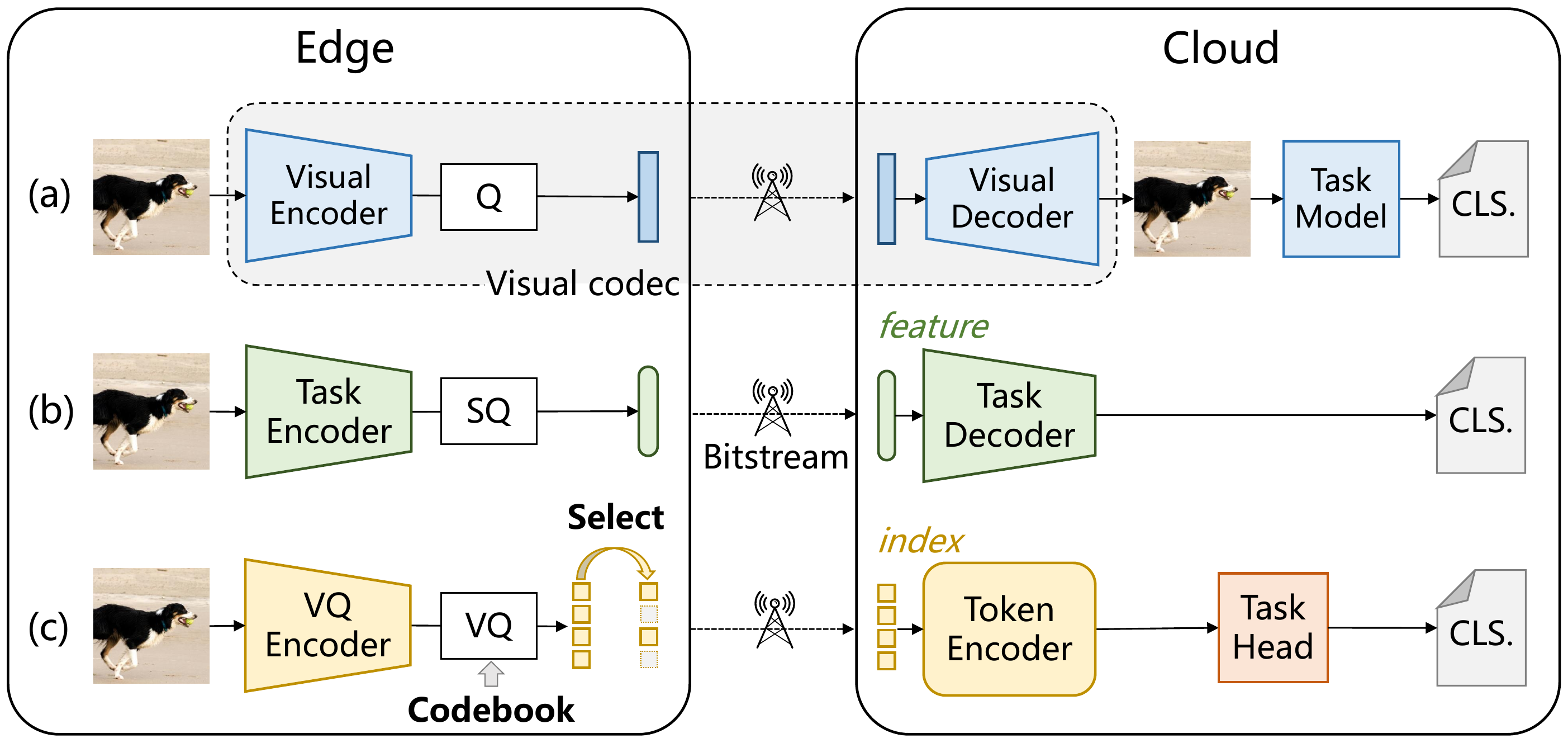}
    \caption{Comparison of edge-cloud collaboration paradigms. (a) Image compression with pixel reconstruction; (b) Feature compression for direct inference; (c) Our proposed codebook-based adaptive feature compression framework.}
    \label{fig:intro}
\end{figure}

A straightforward solution is to deploy the encoder and decoder of visual codecs on the edge and cloud, respectively, and to perform analysis on the reconstructed images, as illustrated in Fig.~\ref{fig:intro} (a). In this pipeline, both traditional rule-based~\cite{bross2021overview, bpg} and learning-based~\cite{minnen2020channel, he2022elic} codecs can be employed. Generally, distortion in the reconstructed images negatively impacts analysis performance. Some approaches~\cite{chen2023transtic, liu2023composable, li2024image} introduce adapters and task constraints to fine-tune the codecs to alleviate this issue. Nonetheless, the codecs are originally designed for human perception instead of machine analysis, inevitably preserving many redundant details in the bitstream. Notably, these methods suffer from poor reconstruction quality under low-bitrate conditions, significantly undermining analysis performance.

\begin{figure*}[t]
    \centering
    \includegraphics[width=0.925\linewidth]{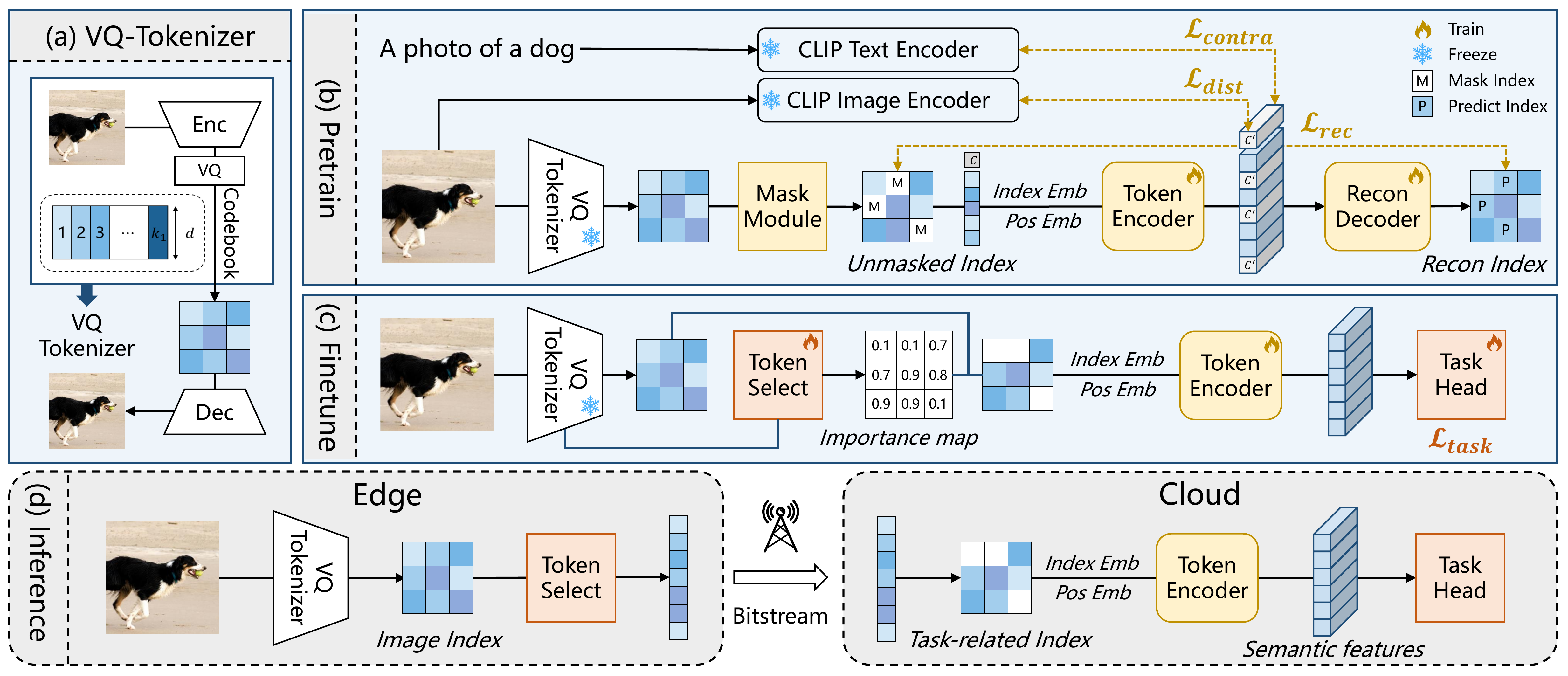}
    \vspace{-2mm}
    \caption{Overview of our Codebook-based Adaptive Feature Compression framework with Semantic Enhancement (CAFC-SE).}
    \label{fig:method}
    \vspace{-2mm}
\end{figure*}

Alternatively, several studies~\cite{matsubara2022supervised,duan2022efficient,hossain2023flexible} compress the intermediate features and conduct analysis directly on the decoded features, as shown in Fig.~\ref{fig:intro} (b). These methods utilize scalar quantization (SQ) with entropy models to encode features and jointly optimize task accuracy and compression rate. For example, Entropic Student~\cite{matsubara2022supervised} splits ResNet into an edge-end for feature compression and a cloud-end for analysis. Despite the progress, the analytical accuracy of these methods is significantly compromised under low-bitrate conditions. This is because SQ quantizes each feature dimension independently, and the entropy model learns the distribution to concentrate on a few high-frequency symbols, which causes the loss of crucial semantic details. Besides, joint optimization~\cite{balle2016end} also tends to reduce the bitrate at the expense of task accuracy under low-bitrate conditions.

In this paper, we propose a Codebook-based Adaptive Feature Compression framework with Semantic Enhancement for the edge-cloud system, dubbed \textit{CAFC-SE}. It offers an alternative paradigm to existing image coding for machine methods, which is less vulnerable to low-bitrate conditions. As shown in Fig.~\ref{fig:intro} (c), CAFC-SE maps continuous visual features to discrete indices with a codebook at the edge via Vector Quantization (VQ) and selectively transmits them to the cloud. Compared with transmitting high-dimensional features from SQ, our method transmits discrete indices from VQ indicating the nearest visual primitives, \ie~codevectors. Hence, CAFC-SE can preserve more informative visual patterns for accurate analysis under low-bitrate conditions.

Technically, CAFC-SE is built upon a lightweight VQ tokenizer to accommodate the resource-limited edge device. Based on the tokenizer, we introduce a semantic-guided masked token modeling approach to pretrain a token encoder to transfer the discrete representation to the semantic feature on the cloud. Furthermore, we design a token selection mechanism to filter out redundant tokens for machine analysis on the edge side, thereby enabling variable-rate feature compression. Extensive experiments demonstrate the superiority of our method in terms of both rate and accuracy.

\section{Proposed Method}
\label{sec:method}
\subsection{Preliminaries}

Vector Quantization (VQ) tokenizers, typically consisting of a visual encoder and a codebook $C = \left \{ c_i \right \} _{i=1}^{N}$, first continuous features from images and then convert them into discrete indices via VQ, providing a compact discrete representation of images~\cite{esser2021taming}. Specifically, given an input image $x \in \mathbb{R} ^{H\times W \times C} $, the encoder maps it into a latent representation $Z \in \mathbb{R} ^{\frac{H}{n}\times \frac{W}{n} \times C} $, where $n$ is the downsample factor. Each vector in $Z$ is subsequently replaced by the nearest codeword from the codebook. Hence, an image can be represented by a discrete index map $I\in \mathbb{R} ^{\frac{H}{n}\times \frac{W}{n}}$, in which each value is the index of the nearest codeword.

Generally, the encoder and codebook are jointly optimized with a decoder on the image reconstruction task, as shown in Fig.~\ref{fig:method} (a). Hence, they learn the discrete representation of the latent space that retains most visual patterns of the images. Despite the impressive results on image generation, discrete representations exhibit inferior performance when directly used for discriminative analysis. This can be attributed to the quantization noise, which increases the difficulty of accurately interpreting semantic information~\cite{peng2022beit,wang2024image}. Nevertheless, the index maps retain essential visual patterns from the input image, thereby providing a technical foundation for developing end-cloud systems under low-bitrate constraints.

\vspace{-2mm}
\subsection{Overview of our feature compression framework}
We first briefly describe our Codebook-based Adaptive Feature Compression framework with Semantic Enhancement (CAFC-SE), designed for low-bitrate image coding for machine. Fig.~\ref{fig:method} depicts the architecture, as well as the training and inference procedures. We construct CAFC-SE based on the VQ tokenizer, whose structure is illustrated in Fig.~\ref{fig:method} (a). Specifically, we first \textit{pretrain} a token encoder to interpret the discrete tokens of the input image into semantic features. As shown in Fig.~\ref{fig:method} (b), we adopt semantic-guided masked token modeling to pretrain the token encoder jointly with a reconstruction decoder, where a CLIP model is used to provide semantic guidance. After pretraining, we remove the reconstruction decoder and introduce a token selection module to filter out less important tokens, which enables variable-rate feature compression. As shown in Fig.~\ref{fig:method} (c), we \textit{finetune} the token encoder with the selection module and the task head on the downstream task.
After pretraining and finetuning, the model can be deployed in the edge-cloud system. As shown in Fig.~\ref{fig:method} (d), the tokenizer and the selection module are deployed on the edge device, while the token encoder and task head are deployed on the cloud server. The sparse indices after selection are transmitted between the edge and cloud. For the tokenizer, we can employ an off-the-shelf lightweight one or train a customized one.

\vspace{-3mm}
\subsection{Semantic-guided masked token modeling}

To accurately extract the semantic information from the discrete tokens for further analysis, we propose a semantic-guided mask token modeling method to train a token encoder. Specifically, we adopt an encoder-decoder architecture and pretrain the token encoder through index map reconstruction, similar to~\cite{li2023mage}. Given the discrete index map $I$, we follow the MAE protocol and randomly mask a subset of tokens with a masking ratio $m \in(0,1)$ and drop out the masked indices. Consequently, we obtain two complementary index sequences: the unmasked index sequence $I_{\Omega}$ and the masked index sequence $I_{\mathcal{M}}$. We leverage the learnable index embeddings $E=\{e_i\}_{i=1}^{N}$ to convert $I_{\Omega}$ into an embedding sequence and add corresponding positional embeddings to the index embeddings. Before feeding this sequence into the token encoder, we also prepend a learnable class token $\left [ C \right ]$ to the sequence. On the output of the token encoder, we fill all masked positions with the encoded class token $\left [ C' \right ]$ corresponding to $\left[ C \right ]$ to recover the original sequence length. The reconstruction decoder ingests the recovered sequence and predicts the masked indices $I_{\mathcal{M}}$. We use the cross-entropy loss as the reconstruction loss:
\begin{equation}
\mathcal{L}_{rec} = -\frac{1}{|\mathcal{M}|} \sum_{i \in \mathcal{M}} \log p_{\theta }\left( {I_i} \mid I_{\Omega}\right),
\end{equation}
where $p_{\theta }\left( {I_i} \mid I_{\Omega}\right)$ is the conditional probability predicted by the reconstruction decoder at position $i$, conditioned on the unmasked tokens $I_\Omega$. $\mathcal{M}$ denotes the set of masked tokens.

The pretext task, \ie~index map reconstruction, does not guarantee that the token encoder can accurately interpret the semantics of the input image. To guide the token encoder in effectively extracting semantic information from the index map, we introduce the vision-language pretrained model, CLIP~\cite{radford2021learning}, as an external semantic teacher to rectify the features output by the token encoder, as shown in Fig.~\ref{fig:method} (b). Concretely, we enforce the embedding of the encoded class token to align with the CLIP embedding. We first employ a two-layer MLP to project this embedding to the dimension of the CLIP embedding and then normalize it, denoted by $z_c$. Next, we impose a cross-modality contrastive loss and a distillation loss on $z_c$. Denoting the image embedding and text embedding of the input image by $z_{img}$ and $z_{text}$, respectively, the distillation loss $\mathcal{L}_{dist}$ and contrastive loss $\mathcal{L}_{contra}$ are formulated as follows:
\begin{equation}
\begin{aligned}
    &\mathcal{L}_{dist} = \phi_{\mathrm{MSE}}(z_{img}, z_c), \\
    &\mathcal{L}_{contra} = \phi_{\mathrm{InfoNCE}}(z_{text}, z_c).
\end{aligned}
\end{equation}
The overall objective $\mathcal{L}$ is
\begin{equation}
    \mathcal{L} = \mathcal{L}_{rec} + \lambda_d \mathcal{L}_{dist} + \lambda_c \mathcal{L}_{contra}.
\end{equation}
where $\lambda_d$ and $\lambda_c$ are the balance weights. $\phi_{\mathrm{MSE}}$ and $\phi_{\mathrm{InfoNCE}}$ denotes the MSE and InfoNCE~\cite{oord2018representation} functions, respectively.

\subsection{Variable-bitrate compression by token selection}
After masked token modeling, the token encoder not only acquires strong semantic representation capabilities but also becomes robust to partial token loss. This property allows us to drop several less important tokens on the edge side, almost without compromising the analysis performance. Hence, we can flexibly set the bitrate by selectively transmitting the important token from edge to cloud according to communication conditions, \ie~variable-bitrate compression.

To this end, we introduce a token selection module to predict the importance score of each discrete token and accordingly drop out the less important tokens, as shown in Fig.~\ref{fig:method} (b).
Specifically, this module takes as input the pre-quantization feature $Z$. It is designed as a lightweight architecture, consisting of a 3×3 depthwise convolutional layer and two projection layers. The depthwise convolutional layer aggregates local contextual information for each token, while the projection layers integrate cross-channel information and generate an importance map $S$. With the importance map, we select tokens with the Top-$K$ score and transmit their indices, denoted by $\mathcal{I}_t$, from edge to cloud. Before transmission, these indices are sorted by spatial position. In addition to the indices, a binary mask is also transmitted to indicate whether the token at each position has been dropped at the edge device. This enables the cloud server to assign the correct positional encoding to each received token.

The token selection module is optimized with the token encoder and task head at the finetuning phase. During finetuning, $K$ is randomly sampled for each training iteration, so that the ability for variable-rate compression can be enhanced.

\section{Experiment}
\label{sec:experiment}
\subsection{Experiment settings}

\textbf{Dataset and metrics.} We evaluate our method on the ImageNet-1K~\cite{deng2009imagenet} dataset for image classification. The input image resolution is resized to 256 $\times$ 256. We report the Top-1 accuracy of the classifier predictions and quantify the compression rate by bits-per-pixel (bpp). In our framework, bpp is related to the number of indices $K$ transmitted over the edge-cloud system and the codebook size.

\noindent \textbf{Baselines.} To verify the effectiveness of our CAFC-SE, we compare it with two edge-cloud collaboration paradigms and focus on classification performance under low-bitrate conditions. \emph{Image Coding} (IC) methods: images are encoded at the edge and reconstructed at the cloud for classification. Baselines include the hand-crafted image codec BPG~\cite{bpg} and learning-based image compression model TransTIC~\cite{chen2023transtic}. \emph{Feature coding} (FC) methods: the edge extracts and compresses intermediate features, which are recovered and directly used for inference in the cloud without pixel reconstruction. Baselines include RAC~\cite{duan2022efficient} and FVR~\cite{hossain2023flexible}. These baseline typically employ a ResNet50(-like) model as the backbone.

\noindent \textbf{Implementation details.} We use the pretrained VQGAN~\cite{chang2022maskgit} model as the VQ tokenizer in our framework. During pretraining, we adopt CLIP~\cite{radford2021learning} as the semantic teacher to guide the masked token modeling and utilize ViT-B~\cite{dosovitskiy2021vit} as the token encoder for classification. During finetuning, we randomly sample the number of task-related tokens $K$. For inference, we evaluate the classification performance of our model by setting $K$ to 256, 225, 196, and 169.

\vspace{-3mm}

\begin{figure}[t!]
    \centering
    \includegraphics[width=1\linewidth]{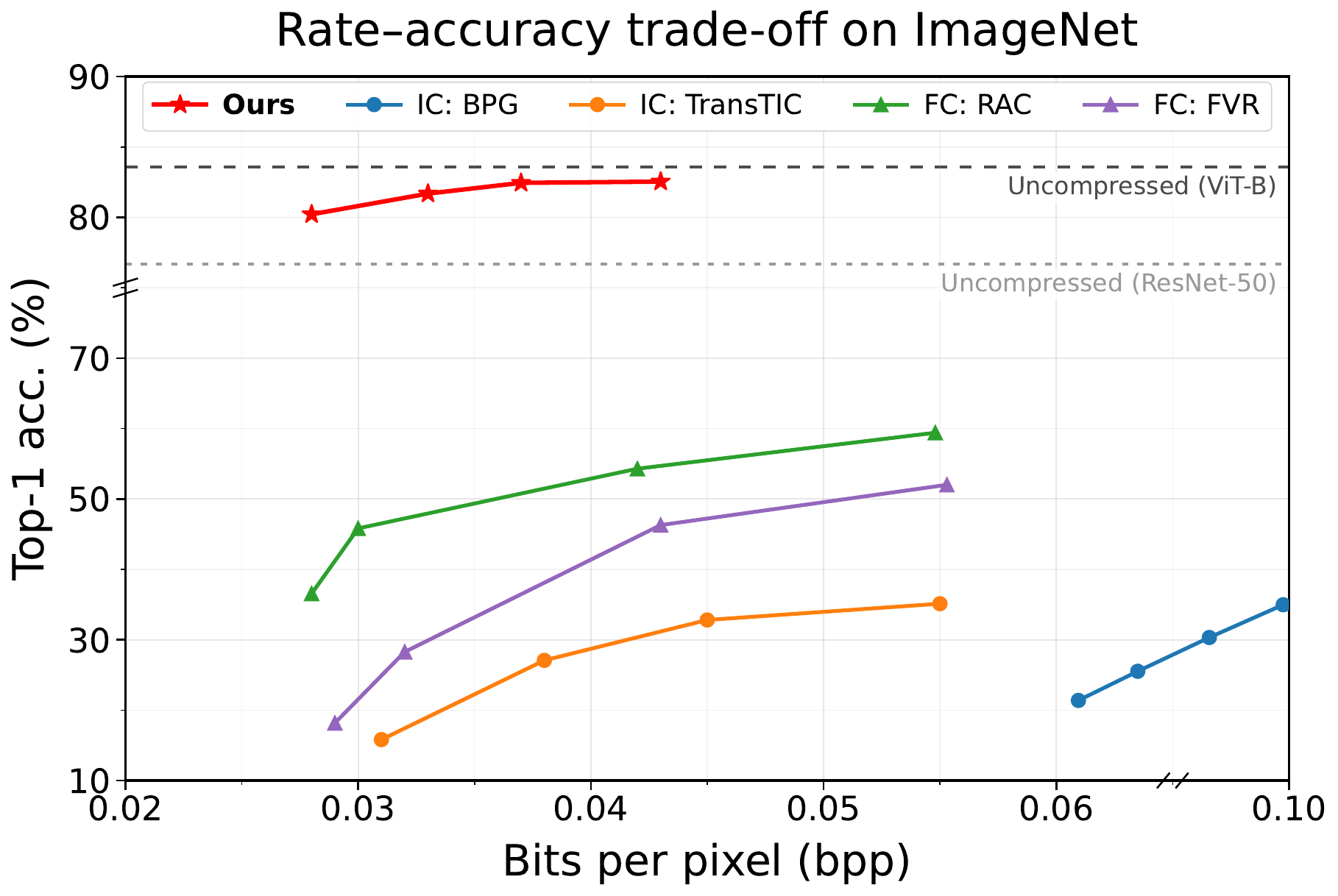}
\vspace{-6mm}
    \caption{Rate-accuracy performance comparison with CAFC-SE and baselines under low-bitrate conditions.}
\vspace{-3mm}
\label{fig:experiment}
\end{figure}

\subsection{Main results}

We compare our CAFC-SE method with the above-mentioned baselines on the classification task under low-bitrate conditions, including two image coding methods (BPG~\cite{bpg} and TransTIC~\cite{chen2023transtic}) and two feature coding methods (RAC~\cite{duan2022efficient} and FVR~\cite{hossain2023flexible}). Fig.~\ref{fig:experiment} illustrates the bpp-acc plot of these methods. Our proposed method CAFC-SE achieves better
 classification accuracy at lower bitrates ($\textless$0.1 bpp). Constrained by scalar quantization and joint rate-distortion optimization, all baselines fail to achieve an effective trade-off between compression rate and task performance, resulting in significant accuracy drops under low-bitrate conditions. In contrast, CAFC-SE pretrains a robust token encoder by semantic-guided masked token modeling and finetunes it with a token selection module for variable-rate compression. Compared with the uncompressed ViT-B, CAFC-SE incurs minor accuracy loss and degrades smoothly across bitrates. 

\subsection{Ablation studies}
We first conduct ablation study experiments to assess the effect of Semantic Enhancement (SE) during pretraining. Hence, we evaluate linear-probing (LP)  classification performance on ImageNet-1K. Table~\ref{tab:main_discrete_compare} reports that our method achieves higher accuracy than w/o SE,  which highlights the effectiveness of SE. Furthermore, we benchmark the edge-cloud paradigms of Fig.~\ref{fig:intro} (a) and (b) with the shared Discrete Features (DF), as shown in Table~\ref{tab:main_discrete_compare}. In paradigm (a), we append a strong decoder and a pretrained classifier after the DF. It achieves competitive classification performance. In paradigm (b), we directly feed discrete features into a ResNet50-like backbone for classification. It obtains inferior performance due to the distribution mismatch.

\begin{table}[t!]
\centering
\captionsetup{skip=4pt}
\setlength{\tabcolsep}{4pt}
\renewcommand{\arraystretch}{0.9}
\caption{Ablation studies of Semantic Enhancement (SE) and edge-cloud paradigms with shared Discrete Features (DF).}
\begin{footnotesize}
\begin{tabular}{lcc}
\toprule
\textbf{Method} & \textbf{Top1-Acc (\%)} & \textbf{$\Delta$ vs Ours (\%)} \\
\midrule
Ours (LP)           & \textbf{78.61} & ---    \\
Ours (LP) w/o SE      & 74.68          & -3.93  \\
\midrule
(a) DF w/ rec   & 72.47          & -6.14  \\
(b) DF w/ conv      & 56.43          & -22.18 \\      
\bottomrule
\end{tabular}
\end{footnotesize}
\label{tab:main_discrete_compare}
\end{table}

\begin{table}[t!]
\centering
\captionsetup{skip=4pt}
\setlength{\tabcolsep}{4pt}
\renewcommand{\arraystretch}{1.05}
\caption{Comparison of Top-1 accuracy at matched bpp with fixed-bitrate and variable-bitrate finetuning.}
\begin{footnotesize} 
% \begin{small}
\begin{tabular}{lcc}
\toprule
\textbf{Test} $K$   & \textbf{Fixed-Finetuning} & \textbf{Variable-Finetuning} \\
\midrule
256 & 82.71\%  & 82.54\% \\
225 & 80.59\% & 82.45\% \\
196 & 79.77\% & 81.69\% \\
169 & 77.16\% & 80.22\% \\
% $K_4$ & $b_4$ & [A$_4$] & [V$_4$] & [V$_4$--A$_4$] \\
\bottomrule
\end{tabular}
% \end{small}
\end{footnotesize}
\label{tab:var_vs_fixK}
\vspace{-3mm}
\end{table}

During finetuning, $K$, which determines the compression rate, can be set to a fixed value (Fixed-Funetuning) or randomly sampled from a range (Variable-Funetuning). We also conduct experiment to analyze the effect of the two strategies. Table~\ref{tab:var_vs_fixK} presents the results of the two strategies on the same testing compression rate ($K$). We observe that variable finetuning achieves slightly higher accuracy at low bitrates than fixed finetuning. The gains arise from random sampling, which improves the generalization ability of the model. Consequently, CAFC-SE can infer across multiple bitrates without retraining and is well-suited for edge-cloud deployment.

\section{conclusion}
\label{sec:conclusion}

In this work, we proposed CAFC-SE, a codebook-based adaptive feature compression framework with semantic enhancement for edge-cloud systems.
CAFC-SE employs a VQ tokenizer to map images into discrete indices, which serve as a compact representation well suited for low-bitrate. We introduce a semantic-guided masked token modeling approach to pretrain a robust token encoder and design a token selection mechanism for variable-rate feature compression. Extensive experiments demonstrate the effectiveness of our method.

\newpage
\bibliographystyle{IEEEbib}
\bibliography{strings,refs}

\end{document}